\documentclass{article}

\usepackage{arxiv}

\usepackage{graphicx}%
\usepackage{multirow}%
\usepackage{amsmath,amssymb,amsfonts}%
\usepackage{amsthm}%
\usepackage{mathrsfs}%
\usepackage[title]{appendix}%
\usepackage{xcolor}%
\usepackage{textcomp}%
\usepackage{manyfoot}%
\usepackage{booktabs}%
\usepackage{algorithm}%
\usepackage{algorithmicx}%
\usepackage{algpseudocode}%
\usepackage{listings}%
\usepackage{hyperref}
\usepackage{url}
\usepackage{longtable}
\usepackage{array}
\usepackage{sidecap}

\title{Selecting a classification performance measure: matching
       the measure to the problem}

\author{\href{https://orcid.org/0000-0002-4649-5622}
  {\includegraphics[scale=0.06]{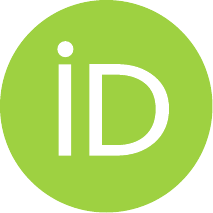}\hspace{1mm}
  David J.~Hand} $^1$ \\
  \texttt{d.j.hand@imperial.ac.uk} \\[3mm]
  $^1$ Department of Mathematics \\
  Imperial College London \\
  London, SW7 2AZ, UK \\
  \And
  \href{https://orcid.org/0000-0003-3435-2015}
  {\includegraphics[scale=0.06]{orcid.pdf}\hspace{1mm}
  Peter Christen} $^{2,3}$ \\
  \texttt{peter.christen@ed.ac.uk} \\[3mm]
  $^2$ Scottish Centre for Administrative \\
  Data Research, University of Edinburgh \\
  Edinburgh, EH8 9BT, UK \\
  \And
  \href{https://orcid.org/0009-0000-5844-0111}
  {\includegraphics[scale=0.06]{orcid.pdf}\hspace{1mm}
  Sumayya Ziyad} $^3$ \\
  \texttt{sumayya.ziyad@anu.edu.au} \\[3mm]
  $^3$ School of Computing \\
  Australian National University \\
  Canberra, 2600, ACT, Australia \\
}

\begin{document}
\maketitle

\begin{abstract}
The problem of identifying to which of a given set of
classes objects belong is ubiquitous, occurring in many research
domains and application areas, including medical diagnosis, financial
decision making, online commerce, and national security. But such
assignments are rarely completely perfect, and classification errors
occur. This means it is necessary to compare classification methods
and algorithms to decide which is ``best'' for any particular
problem. However, just as there are many different classification
methods, so there are many different ways of measuring their
performance. It is thus vital to choose a measure of performance
which matches the aims of the research or application. This paper is
a contribution to the growing literature on the relative merits of
different performance measures. Its particular focus is the critical
importance of matching the properties of the measure to the aims
for which the classification is being made.
\end{abstract}

% --------------------------------------------------------------------




% ====================================================================

\section{Introduction}
\label{sec:intro}

Supervised classification challenges are ubiquitous. They arise in
medical diagnosis, speech and image classification, fraud detection,
personnel selection, astronomical identification, document retrieval,
and an unlimited number of other domains. The generic form of such
problems is as follows:

\begin{enumerate}
\item We have a set of objects, each of which belongs to a known
  class, and for each of which we know its vector of descriptive
  characteristics. This set is often called a training or design
  set.
  \smallskip
  
\item We use this data as the basis for an algorithm, function,
  method, rule sets, or model (we shall henceforth use the term
  \emph{classifier}) which will enable us to classify new objects,
  which have known descriptor vectors but unknown classes, to their
  correct class.
\end{enumerate}

Many special variants of this general problem have also been
explored, and a large variety of names have been given to this
problem across a diverse range of research domains and application
areas. Matching the huge number of applications, a very large number
of different classification methods have been developed. These
include linear discriminant analysis, quadratic discriminant
analysis, naive Bayes, regularised discriminant analysis, logistic
regression, SIMCA, DASCO, logistic regression, perceptrons, neural
networks, deep learning, support vector machines, tree classifiers,
random forests, nearest neighbour methods, Parzen kernel methods,
Gaussian processes, quantile classification, and others
(Fern\'andez-Delgado \emph{et al.} evaluated ``179 classifiers
arising from 17 families''~\cite{Fer14}). However, in most
real-life (as opposed to artificial-world) applications, perfect
classification is impossible: the vector of descriptive
characteristics does not contain sufficient information to perfectly
separate the classes and therefore some cases will be misclassified.
This means that some classification methods are better than others
in any particular problem, and this prompts the obvious question:
which of the many available methods is best? The answer depends on
how one measures classifier performance. The key to choosing the
right measure is to identify what particular aspects of performance
are important for the problem -- as Hand (2001) put it, ``to ensure
that the measure chosen matches the objectives''~\cite{Han01}.

The aim of this paper is to provide a checklist of \emph{properties}
of \emph{problems} and \emph{performance measures} so that an
appropriate choice can be made. This is given in
Table~\ref{tab:properties}. Our work was in part motivated by the
observation that all too often researchers choose a performance
measure on the grounds that ``it was widely used'' or ``others in
this particular research or application domain use it'', regardless
of whether or not it is appropriate. It perhaps goes without saying
that an inappropriate measure of performance can lead to misleading
results. One would not assess the effectiveness of a dietary
regime using a measure of change of height. Likewise, in other
contexts, a poor choice of performance measure result in choosing
a classification method which leads to excessive misdiagnosis, poor
financial decisions, mismatch of personnel to tasks, and worse.

To illustrate the need to choose the right performance measure, we
begin with five motivating examples\footnote{Code for the figures
in Examples 4 and 5 is available from: 
\url{https://github.com/SumayyaZiyad/perf-measures}}.
%{\emph{https://github.com/SumayyaZiyad/perf-measures}}}.
\smallskip

\emph{Example 1}: Perhaps the most popular performance measure is
error rate or misclassification rate (discussed in detail in
Section~\ref{sec:measures}) -- the overall proportion of objects
misclassified by a classifier. In a meta-analysis of empirical
studies of the performance of classification rules, Jamain, page
50, reported that some 97\% of performance results reported
were error rates~\cite{Jam04}. See also Table 1 of Martin (2013)
for a striking illustration of this~\cite{Mar13}. However, a
misclassification rate as low as 1\% is hardly impressive if 99\%
of the objects belong to one of the classes: one could achieve the
same 1\% misclassification rate without the classifier (although,
of course, different objects would be misclassified).
\smallskip

\emph{Example 2}: Continuing with error rate, consider
a situation in which \emph{each} vector in the space of descriptive
characteristics represents a set of objects where their majority is
in class 0. For example, suppose that for every such vector, between
45\% and 49\% of people are sick. Then the overall proportion of
people misclassified is easily minimised by assigning every person
to the healthy class. However, it is unlikely that such an approach
will be very useful. Although it gets over 50\% of the
classifications right, it entirely fails to identify the sick people.
Note that this situation is different from that of Example 1. In the
first example, some of the 1\% misclassified could come from each
class -- see Example 3 for an elaboration of this. Since the basic
aim of classification is to separate the classes, simply classifying
everyone as healthy, or no emails as phishing attempts, rather
defeats the objective of the classification exercise.
\smallskip

\begin{figure}[t]
  \begin{minipage}[c]{0.44\textwidth}
    \includegraphics[width=\textwidth]{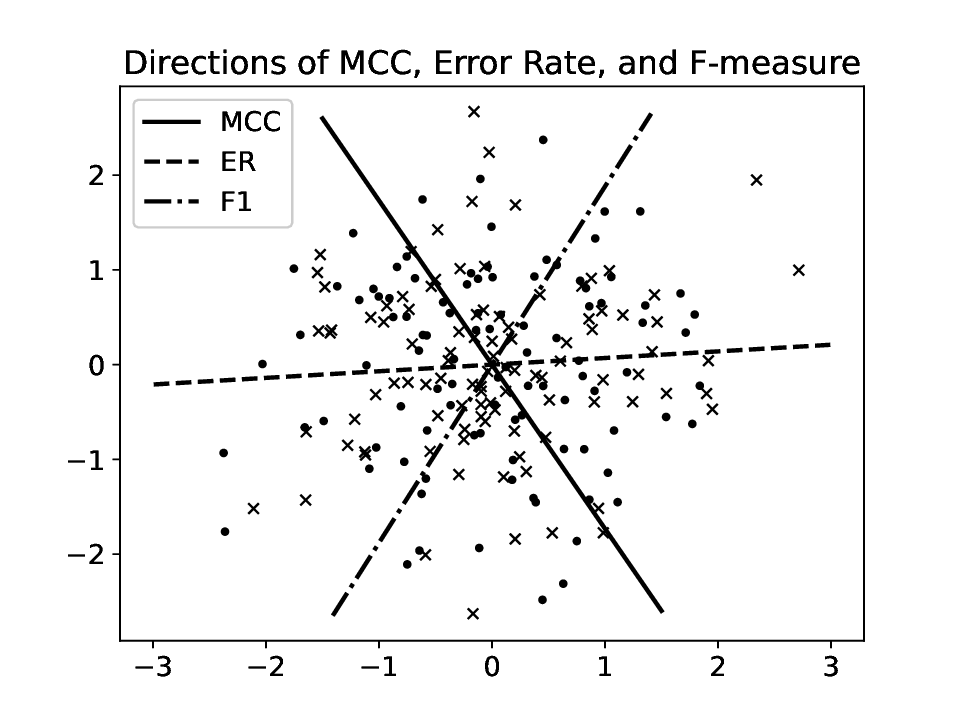}
  \end{minipage}\hfill
  \begin{minipage}[c]{0.47\textwidth}
    \caption{Two simulated sets of 100 data points, and the
    resulting optimal decision boundary for three performance
    measures (Matthews Correlation Coefficient, Error Rate, and
    F-measure, as discussed in Section~\ref{sec:measures}).}
    \label{fig:simdata}
  \end{minipage}
\end{figure}

%\begin{SCfigure}[1][t]
%  \centering
%  \caption{Two simulated sets of 100 data points, and the resulting
%    optimal decision boundary for three performance measures (Matthews Correlation Coefficient, Error Rate, and F-measure, as discussed
%    in Section~\ref{sec:measures}).
%    \label{fig:simdata}}
%  \includegraphics[width=0.49\textwidth]
%    {directions-mcc-fm-er-bw}
%\end{SCfigure}

\emph{Example 3}: Two popular aspects of performance are
\emph{sensitivity} and \emph{specificity} (discussed in
Section~\ref{sec:matrix}). The former is the proportion of class 1
objects (fraudulent credit card transactions, say) which are
correctly categorised by the classification system. The latter is
the proportion of class 0 objects (legitimate transactions) which
are correctly categorised as legitimate by the system. Suppose that
each of these measures has the value of 99\%: that is, 99\% of the
fraudulent cases are correctly classified as fraudulent, and 99\%
of the legitimate cases are correctly classified as legitimate. It
looks as if this is a good classifier: overall it gets 99\% correct.
But now suppose that only about 1 in a 1000 transactions is
fraudulent. Then an elementary calculation shows that 91\% of those
categorised as fraudulent are in fact legitimate. That could be
disastrously poor -- it means that most of the resources spent
investigating apparent frauds are devoted to legitimate transactions,
while also having a detrimental impact on customer relations.
\smallskip

\emph{Example 4}: Figure~\ref{fig:simdata} shows a scatterplot for
synthetic data from two classes, along with the optimal linear
decision boundaries corresponding to three different performance
measures -- the Matthews correlation coefficient, error rate, and
the F-measure, all defined in Section~\ref{sec:measures}. As can
be seen, these separating surfaces lie at completely different
angles. This means that the ``best'' classification according to
one metric may be entirely different from the ``best''
classification according to another. It is clear that the choice
of performance measure is critical to arrive at valid and useful
conclusions.
\smallskip

\begin{figure*}[t]
  \centering
  \includegraphics[width=0.48\textwidth]
    {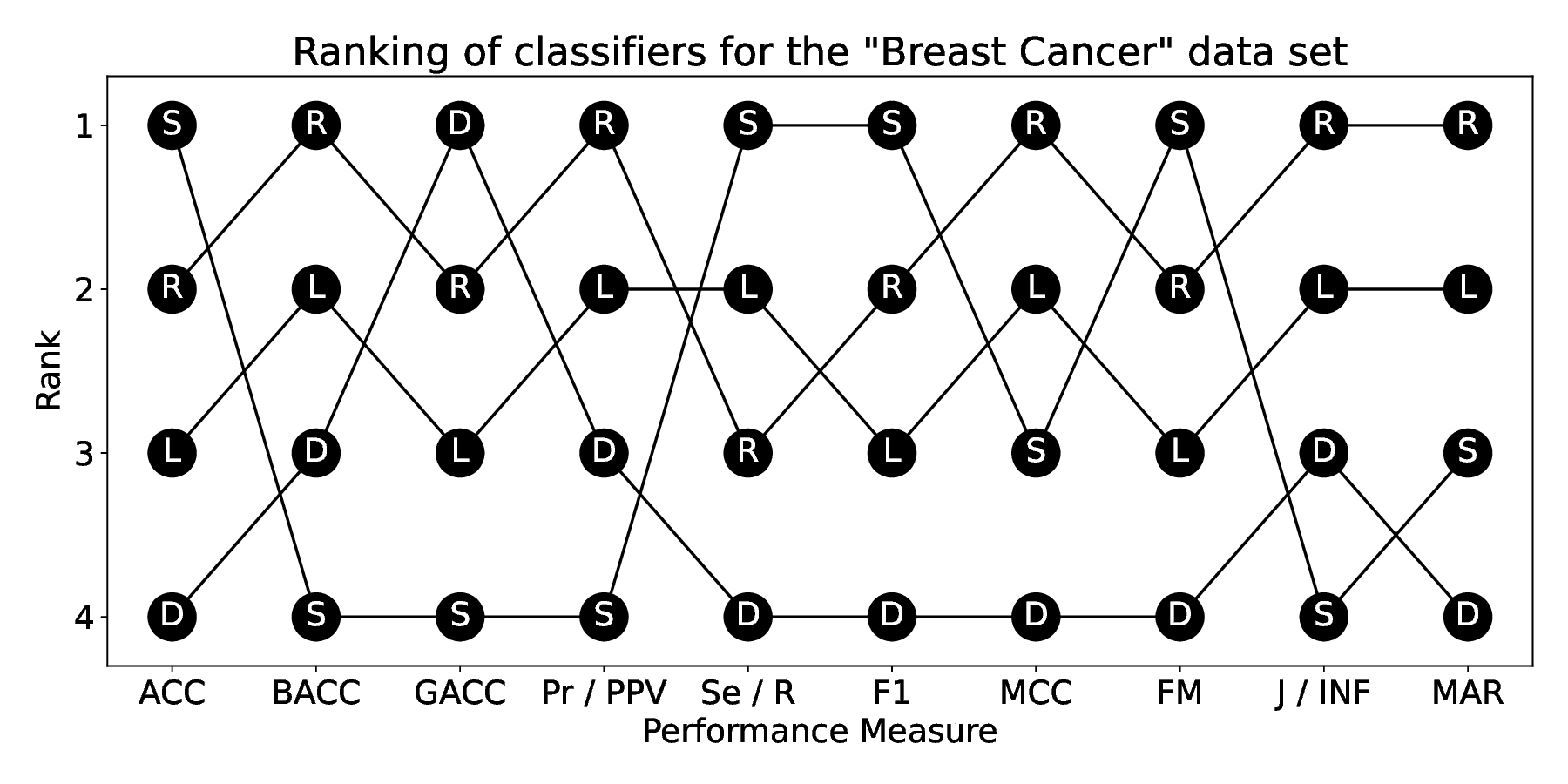} \hfill
  \includegraphics[width=0.48\textwidth]
    {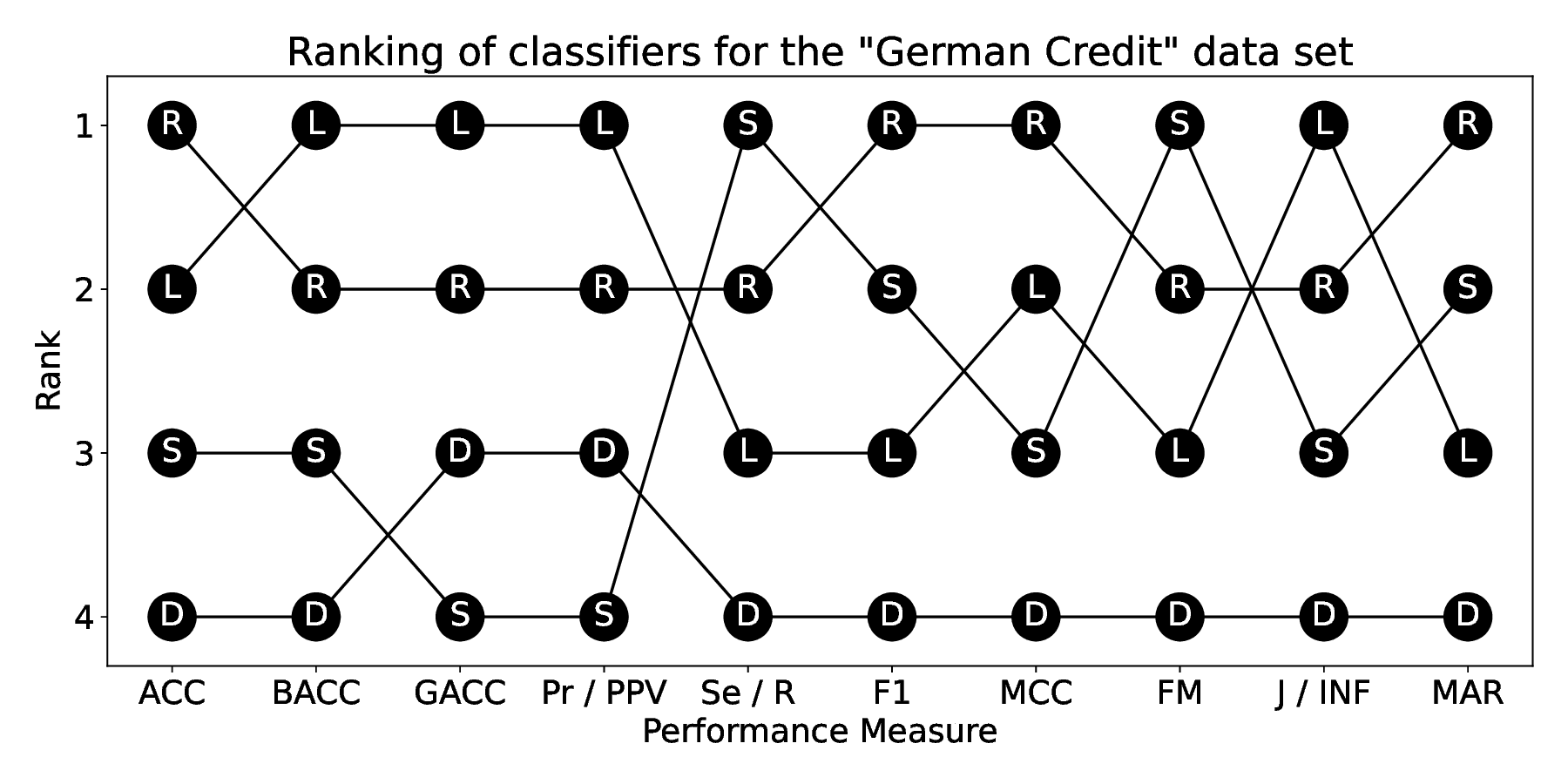}
  \caption{Ranking of classifier performance for two data sets,
    evaluated using ten different performance measures (as defined
    in Section~\ref{sec:measures}). The four classifiers are a
    decision tree (D), logistic regression (L), a random forest
    (R), and a support vector machine (S), as indicated by the
    node labels. \label{fig:ranking}}
\end{figure*}

\emph{Example 5}: Figure~\ref{fig:ranking} shows how different
performance measures can rank classification methods in different
ways. We applied four classification methods to two data sets from
the UCI Machine Learning repository~\cite{Kel24}. The
horizontal axis shows ten different performance measures (all
defined in Section~\ref{sec:measures}), and the vertical axis
shows the rank order of the four classification methods. As can be
seen, the apparent rank of the methods, from 1 (best) to 4 (worst)
can differ radically according to which performance measure is
chosen. Quite clearly, choosing a measure which does not match the
aims could be dramatically misleading.
\medskip

Ideally, a single measure would capture all that we wanted to know
about the performance of a classification method for the problem in
question. If the measure was quantitative (as are all discussed
here), then that would be sufficient to rank the methods and choose
the best for the purpose. However, sometimes it is necessary to take
multiple aspects into account. This can be achieved in two ways:

\begin{enumerate}
\item To use a profile of several performance measures. This does
  carry various risks, including a non-complete ordering
  (classification methods may win on some measures but lose on
  others, as Figure~\ref{fig:ranking} shows) and non-transitivity
  (classifier A beats B and B beats C but C beats A).
  \smallskip

\item Combine the scores of different measures using a numerical
  procedure such as a weighted sum. Of course, this is equivalent
  to defining a new univariate measure.
\end{enumerate}

There are important aspects of choice of performance measure that we
do not attempt to cover in this paper. They include the following.

Firstly, since this paper is concerned with \emph{which measure is
appropriate for the task in hand}, we do not discuss the estimation
or statistical properties of the measures. In particular, for
example, we do not discuss how to avoid the bias arising from
estimating performance using the training data on which the
classifier was built. This has been discussed extensively in the
literature (see, for example, Hand~\cite{Han81},
McLachlan~\cite{Mcl92}, and Hastie \emph{et
al.}~\cite{Has09}). Neither do we discuss statistical tests
of performance measures. Statistical tests can clearly be extremely
important: it is one thing establishing that classifier A scores
better than classifier B on a chosen performance measure and a
given finite test set, but one also needs to know if the difference
could easily have arisen by chance if there was no difference or if,
in fact, classifier B was better than A. Despite its importance,
the issue is not one concerning the conceptual match of the measure
to the objectives, so not appropriate for our discussion here.

Having said that, we note that there are subtleties associated with
statistical testing of measures which are often overlooked. These
include issues of multiple comparisons when several classifiers are
being compared, of dependence between the statistical tests (if the
same test set is to be used for all of the classifiers being
compared), of the sampling scheme through which the test set was
chosen (e.g. randomly sampled from the population or sampling
stratified by the classes?), of how to adjust a statistical test
if the class sizes in the population differ from those in the test
set, and other aspects. Some discussion of testing in this context
is given by Stapor~\cite{Sta17} and Stapor \emph{et al.}~\cite{Sta21}.

A related aspect which we do not discuss is what happens if a
measure cannot be calculated. For example, if both numerator and
denominator in some ratio in a performance measure are zero. These
aspects are important in practical calculation of measures, since
they can cause software to crash (or, worse, give misleading values
without the user being aware of it), but they are not aspects of
the conceptual meaning of the measures and so not appropriate to
discuss in this paper.

Secondly, since we are solely concerned with \emph{classification}
outcome, we do not discuss goodness of calibration of classifier
scores -- how well aligned estimated probabilities of belonging to
each class are to the true probabilities~\cite{Fil23}. This
means we do not discuss measures such as the Brier
score~\cite{Bri50}, which are not directly concerned with
\emph{classification} outcome.

Thirdly, we suppose that only \emph{crisp} performance measures
are being evaluated. The term ``crisp'' is due to Berrar
(2018)~\cite{Ber18}, although he applies it to the classifiers
rather than the measures, contrasting it with ``ranking''
classification methods. Sebastiani~\cite{Seb15} describes them as
``hard'' classifications. A crisp measure is based on both (i) a
classification method which produces scores for each object
\emph{and also} (ii) \emph{a threshold with which the scores are
compared}, as described below. This means that, in particular, we
do not consider measures such as the area under the ROC (Receiver
Operating Characteristic) curve, the area under the precision-recall
(precision-sensitivity) curve, or the H-measure, all of which
average over a range of possible choices of threshold or which use
only rank order information (see, for example, Hand~\cite{Han09}
and Hand and Anagnostopoulos~\cite{Han22}). Although these measures
may be used to compare classifiers, the resulting order of merit may
not match what happens in practice when a particular threshold is
necessarily adopted or if only a small range of thresholds is of
practical relevance~\cite{Mal12}. Such measures have particular
application in situations where the threshold is unknown at the time
a classification method must be chosen.

Likewise, our restriction to crisp measures means we have not
considered measures such as \emph{minimum} error rate or the
Kolmogorov-Smirnov (KS) statistic. These are really particular
values of other measures, using the classification threshold which
optimises those measures.

Fourthly, we are here concerned with \emph{conditional}
classification performance. That means performance of a particular
classifier, with particular numerical values for its parameters,
constructed from a particular training set. This is contrasted
with \emph{unconditional} performance, which is the performance of
an algorithm over possible training sets which could have been
chosen from the population of concern. Rankings of classifiers
produced by conditional and unconditional methods could be very
different, since the conditional measure captures the
idiosyncrasies of a particular training set. Unconditional
performance is useful in deciding what classification method might
be a good choice in future, whereas conditional performance tells
us how good a particular classifier is.

Fifthly, we restrict ourselves to the most important special case:
binary classification methods, where all objects are classified
into one of two classes, labelled 0 or 1. For a review of
multiclass metrics see Grandini \emph{et al.}~\cite{Gra20}.

And sixthly, we note that we are not concerned in this paper with
performance measures that we might term ``non-accuracy-based''
measures. These are very context-dependent and relate to high level
aspects of performance of classification methods. Examples of such
measures are:

\begin{itemize}
\item \emph{Speed of updating}: How quickly a classification method
  can be updated. This is important in non-stationary domains which
  change at a high frequency, such as spam detection.
  \smallskip

\item \emph{Speed of classification}: How quickly a new object can
  be classified. For example, in credit card fraud detection we
  need to classify a transaction as fraudulent or legitimate as it
  is being made.
  \smallskip

\item \emph{Handle large data sets}: How well a classification
  method can handle large data sets. For example, in particle
  physics.
  \smallskip

\item \emph{Handle streaming data}: How well a classification method
  can cope with data which are not simultaneously initially
  available, but are presented as a data stream. For example, in
  telemetry.
  \smallskip

\item \emph{High dimensional problems}: Problems with small numbers
  of classified cases but descriptor vectors with large number of
  dimensions (so-called ``small-n-large-p'' problems). For example,
  some genomics problems.
  \smallskip

\item \emph{Handle missing data}: In some applications, missing data
  are common, so it can be important how effective the method is in
  coping with incomplete data.
  \smallskip

\item \emph{Interpretability of a classification method}: How
  interpretable is a method's approach to classification. This has
  become a hot topic with the advent of methods such as deep
  learning, and increasing interest in fairness and bias in
  artificial intelligence.
  \smallskip

\item \emph{Ease of use}: There is no better description of this
  aspect than that given by Duin (1996, page 535), who wrote: ``in
  comparing classifiers one should realize that some classifiers
  are valuable because they are heavily parameterized and thereby
  offer a trained analyst a large flexibility in integrating his
  problem knowledge in the classification procedure. Other
  classifiers, on the contrary, are very valuable because they are
  entirely automatic and do not demand any user parameter
  adjustment. As a consequence they can be used by
  anybody.''~\cite{Dui96}.
\end{itemize}

There are a great many papers which explore these other aspects of
performance, including Duin~\cite{Dui96}, Salzberg~\cite{Sal97},
and Hand~\cite{Han06}.

Section~\ref{sec:matrix} of the paper describes the mathematical
setup and notation. Section~\ref{sec:properties} then discusses
properties of performance measures. This section, in particular,
characterises the distinction between this paper and other reviews.
We identify two distinct types of properties of performance
measures (beyond the ``accuracy / non-accuracy'' distinction already
noted): (i) structural properties; and (ii) properties related to
aspects of the research problem or the application domain. Other
reviews typically elide the distinction. We consider the second
type critical in ensuring that the performance measure is measuring
the aspects that matter most to the researcher, and this is our
focus. Section~\ref{sec:measures} lists performance measures,
presenting in Table~\ref{tab:properties} the cross-classification
of measure by properties of the second type discussed in
Section~\ref{sec:properties}. Section~\ref{sec:related} describes
other work on comparing performance measures and
Section~\ref{sec:conclusions} summarises and presents our
conclusions.

% --------------------------------------------------------------------

\begin{figure}[t]
  \centering
  \includegraphics[width=0.55\textwidth]{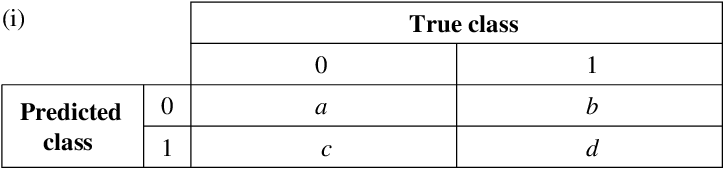}
  \vspace*{4mm}
  
  \includegraphics[width=0.55\textwidth]{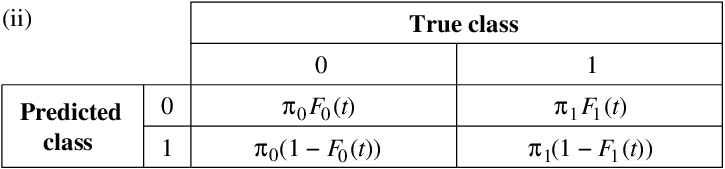}
  \vspace*{4mm}
  
  \includegraphics[width=0.55\textwidth]{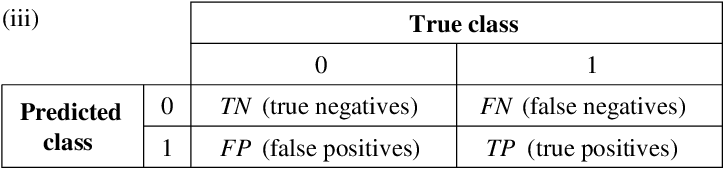}
  
  \caption{Notation for confusion matrix.}
  \label{fig:matrix}
\end{figure}

\section{The confusion matrix}
\label{sec:matrix}

The generic form of a classification method is a function mapping the
descriptive vector of each object to a score on a scale. This score
is compared with a threshold value $t$. Objects with scores larger
than the threshold are classified as belonging to class 1 and other
objects as belonging to class 0. The core elements of any
classification method are thus: (i) a function yielding a
distribution of scores for each of the classes, and (ii) a choice
of threshold value.

When applied to a test set of objects, each with known descriptor
vectors and class labels, this rule yields a \emph{confusion} matrix,
as illustrated in Figure~\ref{fig:matrix} (i), which shows the
number of test set objects belonging to class 0 which are correctly
classified as class 0 (\emph{a}), the number of class 0 objects
incorrectly classified as class 1 (\emph{c}), and so on. We see that
the confusion matrix arising from applying a classifier to a data
set contains four values. Our aim is to reduce those four to a
single number so that different classifiers can be compared. The
wealth of performance measures arises from the fact that this
reduction can be achieved in different ways. However, note that,
for a given test set, the counts $a + c$ and $b + d$ are the same
for all classifiers (that is, for all score distributions and
threshold choices), representing the numbers in each class in the
test set. What differs between classifiers are the splits of
$a + c$ into $a$ and $c$, and $b + d$ into $b$ and $d$. Obviously,
also $n = a + b + c + d$, the total number of objects in the test
set, is fixed. We thus have only two degrees of freedom in the
confusion matrix. Our aim then is to choose which two degrees of
freedom to use to describe the confusion matrix, and how to reduce
the two-dimensional space to a univariate continuum which can be
used to compare matrices (and therefore classifiers). Clearly, it
is possible to express any single measure in terms of any of the
chosen pairs of degrees of freedom.

We can write a generic performance measure defined on
Figure~\ref{fig:matrix} (i) as $M\left( {a,b,c,d} \right)$.

There is an analogous theoretical table, Figure~\ref{fig:matrix}
(ii), in which the threshold is applied to the underlying true
distributions of scores, $F_0(x)$ and $F_1(x)$. Here $\pi_1$
represents the proportion of objects which belong to class 1,
$\pi_0$ the proportion which belong to class 0, $F_0(t)$ the
proportion of class 0 with scores less than $t$, and $F_1(t)$ the
proportion of class 1 with scores less than $t$.

%Often one class (which we label class 1) is regarded as the
%``positive'' class (e.g. in fraud detection, disease screening,
%etc). In that case we can speak of true negative counts (TN), false
%negative counts (FN), false positive counts (FP), and true positive
%counts (TP), as in Figure~\ref{fig:matrix} (iii).

Often one class (which we label class 1) is regarded as the
``positive'' class
%(e.g. in fraud detection, disease screening, and so on).
(e.g. fraudulent cases in fraud detection, patients having an
illness in disease screening, and so on). In that case we can
speak of true negative counts (TN), false negative counts (FN),
false positive counts (FP), and true positive counts (TP), as in
Figure~\ref{fig:matrix} (iii).

As well as the notation $n = a + b + c + d$, and the test set class
proportions $\pi_0 = (a + c)/n$ and $\pi_1 = (b + d) / n$, we
shall denote the test set predicted proportions by $p_0 = (a + b)
/n$ and $p_1 = (c + d) / n$.

Begging the reader's indulgence, for expository convenience in what
follows we shall present the definitions in terms of the notation of
Figure~\ref{fig:matrix} (i). Technically, the resulting values will
merely be estimates of the corresponding measures, based on the
observed cell counts arising from a test set, but it is easier to
see the meaning of a ratio like $2d / (b + c + 2d)$ than one like
%
% \[{{2\pi_{1}\left( {{1 - F}_{1}(t)} \right)}/\left\{ {\pi_{1}F_{1}{(t) + \pi_{0}}{\left( {{1 - F}_{0}(t)} \right) + 2}\pi_{1}\left( {{1 - F}_{1}(t)} \right)} \right\}}.\]
%
\begin{equation}
 \frac{2 \pi_1(1 - F_1(t))}
    {\pi_1 F_1(t) + \pi_0(1 - F_0(t)) + 2 \pi_1 (1 - F_1(t))}.
    \nonumber
\end{equation}

Some important choices for the degrees of freedom are simple ratios
based on the rows and columns of the confusion matrix. For example,
in medicine and epidemiology \emph{sensitivity} and
\emph{specificity} of diagnostic tests (both discussed in
Section~\ref{sec:matrix}) are a popular pair, these being defined
as
\begin{equation}
  \mathit{Se} = d / (b + d) \nonumber
\end{equation}
and
\begin{equation}
  \mathit{Sp} = a / (a + c) \nonumber
\end{equation}

In contrast, in information retrieval, the pair \emph{recall} (the
same as \emph{sensitivity}) and \emph{precision} are widely used,
these being defined as
\begin{equation}
  \mathit{R} = d / (b + d) \nonumber
\end{equation}
and
\begin{equation}
  \mathit{Pr} = d / (c + d) \nonumber
\end{equation}

The reader will see from this that some measures go under more than
one name (e.g.~sensitivity and recall; and precision and
\emph{positive predictive value}), according to the discipline in
which they were developed and are used. Some indeed go under a wide
variety of names. Furthermore, some measures are one-to-one
transformations of others (e.g. \emph{error rate} = 1 -- 
\emph{accuracy}).

Finally, note the fact that each performance measure reduces the four
numbers in the confusion matrix to a single value necessarily implies
that different sets of numbers will result in the same performance
value. For example, for a given test set size, simple
misclassification rate is invariant to how the misclassified points
are distributed across the two kinds of misclassification. That
means that for any performance measure one can define various kinds
of isoeffectiveness curves, with different confusion matrices
yielding the same value.

% --------------------------------------------------------------------

\section{Properties of performance measures}
\label{sec:properties}

As we noted at the end of Section~\ref{sec:intro}, properties of
(``accuracy-based'') performance measures can be divided into two
types, of which the second type is the primary concern of this
paper. The first type are properties, often mathematical properties
(we shall call them \emph{structural properties}), which might be
useful to have and which might aid interpretation, but which do not
reflect constraints one wishes to capture arising from the research
problem or application. This first type may be axioms which one
would like one's measure to satisfy -- the fact that a measure can
take values only in a finite interval, or can only be positive, for
example. These properties are not our concern here and we shall not
discuss them beyond producing the illustrative list immediately
below.

%  - - - - - - - - - - - - - - - - - - - - - - - - - - - - - - - - - -

\subsection{Structural properties of performance measures}
\label{sec:structural}

Structural properties of performance measures, often based on
axioms which are taken to be self-evidently desirable, include:

\begin{itemize}
\item \emph{Direction}: Does ``large'' mean ``good''? For example,
  larger is better for proportion correctly classified but worse for
  error or misclassification rate.
  \smallskip

\item \emph{Maximum}: Is the measure bounded above? For example,
  the proportion correctly classified is necessarily no greater
  than 1.
  %For single degree of freedom (d.o.f.) measures, the maximum
  %indicates that all the relevant objects are correctly classified.
  %For example, for sensitivity, are all true class 1 objects
  %correctly classified?
  \smallskip

\item \emph{Minimum}: Is the measure bounded below? For example,
  the proportion correctly classified is necessarily no less than
  0.
  \smallskip

\item \emph{Interval}: Related to both of the previous two
  properties, is the measure constrained to lie in an interval?
  This is often taken to be the unit interval, $[0,1]$.
  \smallskip

\item \emph{Monotonicity}: Does changing a confusion matrix by
  correctly classifying an incorrectly classified point lead to an
  improvement in the value of the performance measure. In the case
  of binary classification methods, this means that, for performance
  measures in which larger is better
  \begin{equation}
  M(a+1, b, c-1, d) \geq M(a, b, c, d) \nonumber
  \end{equation}
  and
  \begin{equation}
  M(a, b-1, c, d+1) \geq M(a, b, c, d), \nonumber
  \end{equation}
  with the inequalities inverted for measures in which smaller is
  better (such as for error rate).
  \smallskip

\item \emph{Baseline adjusted}: Is the measure adjusted for some
  baseline derived from the confusion matrix? This is distinct from
  situations in which an external baseline is used (e.g. performance
  of a standard treatment or placebo). Different baseline measures
  can be used. For example, Megahed \emph{et al.}~\cite{Meg24}
  describe three baselines which are (i) class label predictions
  being random with equal probabilities of 0 and 1, (ii) class
  label prediction being random with probabilities the same as the
  observed relative class sizes ($(a + c) / n$ and $(b + d) / n$),
  and (iii) all labels predicted to be the label of the larger
  class. Baseline adjustment will generally mean that the maximum
  or minimum of the measure has a fixed value (e.g. 0 or 1).
  \smallskip

\item \emph{Constant baseline}: This property proposed by G\"osgens
  \emph{et al.} (2021)~\cite{Gos21} requires that the scores
  resulting from random assignments of $n$ objects to classes
  should achieve the same expected values, regardless of the number
  of objects assigned to each class. For example, a roughly equal
  distribution of class sizes should achieve the same expected value
  for the performance measure as a markedly unequal distribution
  (excluding the special case in which all objects are assigned to
  one class). It is a property that avoids high measures arising
  purely because (for example) a majority of objects are assigned
  to one class, regardless of the predictive accuracy of a
  classification method.
\end{itemize}

%  - - - - - - - - - - - - - - - - - - - - - - - - - - - - - - - - - -

\subsection{Properties of performance measures arising from the
            research or application aims}
\label{sec:application}

The second type of property of performance measures reflect
constraints imposed by the research aim, and is our concern. The
properties we examine are listed below. We make no claim that this
is comprehensive. Indeed, the scope of application of supervised
learning means that the range of different kinds of problem is
huge, and doubtless other properties arise from research aims we
have not considered. Other researchers have discussed these measures,
but as far as we are aware no-one has highlighted this type of
aspect as of critical importance to the aim of evaluating and
choosing between classification methods. The properties we consider
are:

\begin{itemize}
\item \emph{Classification costs}: The precise structure of the
  costs will depend on the problem. The most common situation is
  for \emph{misclassifications} to carry costs, and for other costs
  to form a common baseline which can be ignored in a performance
  measure~\cite{Dru06}. Because this special case is so important,
  and is by far the situation most commonly adopted, we have
  singled it out. However, more general cost structures do arise
  in special problems, and choosing a measure based on the familiar
  assumed structure could result in misleading results. An example
  arises in plastic card fraud detection, where one may have
  reasonable estimates of various kinds of costs, including the cost
  of failing to detect a fraud, of misclassifying a legitimate
  transaction as fraudulent, and of investigating transactions
  flagged as potentially fraudulent~\cite{Han08}. In some situations
  it is more appropriate to think in terms of simple importance
  weights, rather than costs (such as $F_\beta$, described in
  Section~\ref{sec:measures}), or in terms of losses, benefits,
  utilities, and so on.
  \smallskip

\item \emph{Complete}: Does the measure take account of all four
  cells of the confusion matrix? For example, in information
  retrieval the number of irrelevant documents correctly
  classified as irrelevant may well have no bearing on the
  classification performance, which depends on how many relevant
  documents are identified and how many irrelevant documents are
  mistakenly classified as relevant. Likewise, measures which
  focus on a single degree of freedom (e.g. a ratio of just two
  of the four cells) are not complete.
  \smallskip

\item \emph{Symmetry}: Does the measure treat the two classes on
  an equal footing. In particular, would we get the same numerical
  value if we switched the class labels. For example, fraud
  detection (a transaction is either fraudulent or not) problems
  are generally asymmetric, but a speech recognition system (two
  classes: spoken ``yes'' or ``no'') is generally symmetric. This
  is related to the common cost structure described above, in
  which only misclassifications carry costs.
  \smallskip

\item \emph{Meaning}: Some measures are
  \emph{representational}~\cite{Han04}, implying they have a
  ready interpretation as real-world properties (e.g. as a
  probability -- such as misclassification rate, which is the
  probability that a case is misclassified). Others are purely
  \emph{pragmatic}~\cite{Han04}, meaning they are artificial
  constructs capturing aspects of interest, but in a way
  corresponding to no underlying real property. One example is
  the F-measure. While this has an interpretation as a harmonic
  mean of two probabilities, the result cannot be interpreted as
  the probability of any random event~\cite{Chr23}.
  \smallskip

\item \emph{Balanced:} Is the measure not biased in favour of the
  majority class? For example, balanced accuracy (discussed in
  Section~\ref{sec:measures}) treats the proportions of each class
  misclassified equally, so that it is balanced on imbalanced data
  sets. Example 3 of Section~\ref{sec:intro} demonstrated that a
  classifier which correctly classified the majority of both the
  positive and negative instances might in fact be very imbalanced,
  depending on the aims.
  \smallskip

\item \emph{Ignores correctly classified objects from one class}:
  In some situations not all cell counts of the confusion matrix
  are relevant. In information retrieval, for example, the
  potentially vast number of irrelevant documents that are
  correctly identified as irrelevant should have no bearing on
  the performance. Clearly this property is the complement of the
  completeness property, so we need not have explicitly defined
  this property. However, since it is important to recognise that
  \emph{either possessing the property or not possessing it} might
  be appropriate and desirable for the research aim, we have
  included it.
\end{itemize}

% --------------------------------------------------------------------

\section{Performance measures}
\label{sec:measures}

This section lists performance measures and the different names they
go under. Given the wealth of measures, and the diversity of names
for each, it is likely that we have missed some, and in such cases
we would welcome readers sending us details of any we have omitted
(especially if they are novel measures, and not simple alternative
names for or transformations of existing measures). At the end of
this section we show in Table~\ref{tab:properties} which of the
presented measures fulfil which of the properties we discussed in
Section~\ref{sec:application} above.

For each measure, we give its definition in terms of the confusion
matrix shown in Figure~\ref{fig:matrix} (i), and also properties
relevant to its use. Some of the measures provide single degree of
freedom (d.o.f.) summaries and need to be combined with another
summary to yield an overall measure.

\begin{itemize}
\item \emph{Se}: \emph{Sensitivity}: 
  \begin{equation}
  \mathit{Se} = \frac{d}{b + d}
  \end{equation}
  Also called \emph{Recall (R)}, \emph{Hitrate}, and \emph{True
  Positive Rate}. This treats class 1 as the ``cases'' (e.g.
  instances of a particular illness being diagnosed, frauds, etc),
  and is the proportion of cases correctly classified. As noted
  above, sensitivity by itself is not a complete summary of the
  confusion matrix, and so needs to be combined with another
  d.o.f.~to yield a complete performance measure. This can be done
  in many ways and other measures below are examples.
  \smallskip

\item \emph{Sp}: \emph{Specificity}:
  \begin{equation}
  \mathit{Sp} = \frac{a}{a + c}
  \end{equation}
  Also called \emph{Selectivity}, and \emph{True Negative Rate}.
  This treats class 1 as the ``cases'' and is the proportion of
  \emph{non-cases} correctly classified.
  \smallskip

\item \emph{Pr}: \emph{Precision}:
  \begin{equation}
  \mathit{Pr} = \frac{d}{c + d}
  \end{equation}
  Also called \emph{positive predictive (or predicted) value}
  (\emph{PPV}). This also treats class 1 as the ``cases'' and is
  the proportion of those predicted to be cases which actually are
  cases. Note that the term ``precision'' is also used in the very
  different sense of how well-calibrated are estimates of class 1
  probabilities given by a classifier~\cite{Buj05}.
  \smallskip

\item \emph{FDR}: \emph{False Discovery Rate}:
  \begin{equation}
  \mathit{FDR} = \frac{c}{c + d}
  \end{equation}
  Also called \emph{false positive rate} and \emph{fallout}. It is
  the complement of \emph{Pr}: \emph{FDR} = 1 - \emph{Pr}.
  \smallskip

\item \emph{NPV}: \emph{Negative Predictive (or Predicted) Value}:
  \begin{equation}
  \mathit{NPV} = \frac{a}{a + b}
  \end{equation}
  This also treats class 1 as the ``cases'' and is the proportion
  of those predicted to be \emph{non-cases} which actually
  are non-cases.
  \smallskip

\item \emph{FOR}: \emph{False Omission Rate}:
  \begin{equation}
  \mathit{FOR} = \frac{b}{a + b}
  \end{equation}
  The complement of \emph{NPV}: \emph{FOR} = 1 - \emph{NPV}.
  \smallskip

\item \emph{ACC}: \emph{Accuracy}:
  \begin{equation}
  \mathit{ACC} = \frac{a + d}{n}
  \end{equation}
  This is a simple count of the proportion correctly classified.
  It can be written in various ways as a combination of the single
  d.o.f.s above. For example:
  \begin{equation}
  \mathit{ACC} = \pi_0 \cdot \mathit{Sp} + \pi_1 \cdot \mathit{Se}
  \nonumber
  \end{equation}
%  \smallskip

\item \emph{BACC}: \emph{Balanced Accuracy}:
  \begin{equation}
  \mathit{BACC} = \frac{1}{2} \left(\frac{a}{a + c} + \frac{d}{b + d}
  \right)
  \end{equation}
  This is similar to \emph{ACC} except with equal weights on
  \emph{Sp} and \emph{Se}:
  \begin{equation}
  \mathit{BACC} = \frac{1}{2} \mathit{Sp} + \frac{1}{2}\mathit{Se}
  \nonumber
  \end{equation}
  %\smallskip

\item \emph{GACC}: \emph{Geometric Accuracy}:
  \begin{equation}
  \mathit{GACC} = \sqrt{\left( \frac{d}{b + d} \right)
    \left( \frac{a}{a + c} \right)} = \sqrt{\mathit{Se} \cdot
    \mathit{Sp}}
  \end{equation}
  This is the geometric mean of specificity and sensitivity. Like
  \emph{BACC} it gives equal weight to the two components.
  \smallskip

\item \emph{ER}: \emph{Error Rate}:
  \begin{equation}
  \mathit{ER} = \frac{b + c}{n}
  \end{equation}
  Also called \emph{misclassification rate}. This is a simple count
  of the proportion misclassified -- the complement of
  \emph{ACC}:
  \begin{equation}
  \mathit{ER} = 1 - \mathit{ACC} \nonumber
  \end{equation}
  It can be written as a weighted average of \emph{Se} and
  \emph{Sp}, with the weights being the class sizes:
  \begin{equation}
  \mathit{ER} = \left(\frac{a + c}{n} \right) \left( 1 - \mathit{Sp}
    \right) + \left(\frac{b + d}{n}\right) \left( {1 - \mathit{Se}}
    \right), \nonumber
  \end{equation}
  or
  \begin{equation}
  \mathit{ER} = \pi_0 ( 1 - \mathit{Sp}) + \pi_1
  ( 1 - \mathit{Se}), \nonumber
  \end{equation}
  or, of course, in terms of combinations of other single d.o.f.
  aspects of performance. It is also equal to the \emph{Hamming
  distance} between the true and predicted vectors of class labels
  divided by the test sample size, $n$.

  As we noted above, error rate is the most widely-used measure of
  classifier performance~\cite{Jam04,Mar13}. It has the important
  property that it regards the two kinds of misclassification as
  equally serious. We contend that for most problems this is not
  appropriate, and that normally one kind of misclassification is
  more serious than the reverse. The next measure tackles this.
  \smallskip

\item \emph{WER}$(k)$: \emph{Weighted Error Rate}:
  \begin{equation}
  \mathit{WER}(k) = \frac{k \cdot b + (1 - k) \cdot c}{n}
  \end{equation}
  This is a modification of \emph{ER} in which the relative
  severities of the two kinds of misclassification are in
  proportion $k / (1-k)$: misclassifying a class 1 object is
  $k / (1-k)$ times as serious as misclassifying a class 0 object.
  Or, the cost of misclassifying a class 1 object is $k$ and
  the cost of misclassifying a class 0 object is $(1-k)$.

  Variants of this using benefits instead of costs or taking a
  different baseline can easily be derived (see, for example,
  \emph{net benefit} in Vickers and Elkin~\cite{Vic06} and $M_2$
  in Hand~\cite{Han05}). Note that, in principle, all performance
  measures based on a single confusion matrix can be modified to
  consider costs or benefits similarly to the way in which ER is
  modified to WER.
  \smallskip

\item \emph{K}: \emph{Cohen's Kappa}~\cite{Coh60,Chi21}:
  \begin{equation}
  \mathit{K} = \frac{2 (a \cdot d - b \cdot c)}{(c + d)(a + c) +
               (b + d) (a + b)}
  \end{equation}
  That is
  \begin{equation}
  \mathit{K} = \frac{2 (a \cdot d - b \cdot c) / n^2}
               {p_1 \pi_0 + p_0 \pi_1}. \nonumber
  \end{equation}
  This can alternatively be written as
  \begin{equation}
  \mathit{K} = \left(\frac{a + d}{n} - p_0 \pi_0 - p_1 \pi_1
               \right) / \left( 1 - p_0 \pi_0 - p_1 \pi_1 \right),
               \nonumber
  \end{equation}
  That is $(\mathit{ACC} - p_0 \pi_0 - p_1 \pi_1) / ( 1 - p_0 \pi_0
  - p_1 \pi_1)$. Here, $(p_0 \pi_0 + p_1 \pi_1)$ is the chance
  proportion correctly classified by a classifier which assigns
  objects to classes at random in the proportions in the test set.
  So \emph{K} is the extent to which the proportion correctly
  classified exceeds the chance proportion correct expressed as a
  proportion of the total possible improvement.
  \smallskip

\item \emph{J}: \emph{Youden's J}, also known as the
  Youden index:
  \begin{equation}
  \mathit{J} = \frac{d}{b + d} + \frac{a}{a + c} - 1
  \end{equation}
  That is,
  \begin{equation}
  \mathit{J} = \mathit{Se} + \mathit{Sp} - 1 \nonumber
  \end{equation}
  Also called \emph{Informedness (INF)}~\cite{Pow11} and
  \emph{Net Reclassification Improvement}~\cite{Gu09}, this
  measure combines the two single d.o.f.~measures of Sensitivity
  and Specificity into an overall measure. It can also be thought
  of as the difference between the proportions of class 1 and
  class 0 assigned to class 1. It is also a linear transformation
  of the (unweighted) average of \emph{Se} and \emph{Sp}.
  \smallskip

\item \emph{MAR}: \emph{Markedness} (also called
  \emph{DeltaP}~\cite{Pow11}):
  \begin{equation}
  \mathit{MAR} = \frac{d}{c + d} + \frac{a}{a + b} - 1
  \end{equation}
  That is,
  \begin{equation}
  \mathit{MAR} = \mathit{\Pr} + \mathit{NPV} - 1 \nonumber
  \end{equation}
  %\smallskip

\item \emph{F}$_1$: \emph{F-measure}, often simply called \emph{F}:
  \begin{equation}
  \mathit{F}_1 = \frac{2d}{b + c + 2d}
  \end{equation}
  This is the harmonic mean of \emph{Se} (Sensitivity or Recall)
  and \emph{Pr} (Precision) or
  \begin{equation}
  \mathit{F}_1 = \frac{2}{(\mathit{Se}^{-1} + \mathit{Pr}^{-1})},
    \nonumber
  \end{equation}
  and we see that the value of $a$ does not contribute to the
  measure. This is useful in certain applications -- such as, for
  example, information retrieval -- where one might want an
  arbitrarily large number of correctly classified irrelevant
  documents not to inflate the measure.
  \smallskip

\item \emph{F}$_\beta$: \emph{Weighted \emph{F}$_1$}:
 \begin{equation}
  \mathit{F}_\beta = \left[ \frac{\alpha}{d / (c + d)} + \frac{1 -
    \alpha}{d / (b + d)} \right]^{-1}
  \end{equation}
  That is
 \begin{equation}
  \mathit{F}_\beta = \left[ \frac{\alpha}{Pr} + \frac{1 - \alpha}
    {Se} \right]^{-1} \nonumber
  \end{equation}
  We see that this is a weighted harmonic mean of precision and
  sensitivity (recall), where the relative importance of these two
  components, given by $\alpha$ and $(1-\alpha)$, is specified from
  external considerations~\cite{Chr23}.

  This is sometimes written in the alternative form (where
  $\beta = \sqrt{\alpha^{-1} - 1}$):
 \begin{equation}
  \mathit{F}_\beta = (1 + \beta^2) \frac{\mathit{\Pr} \cdot
    \mathit{Se}} {\beta^2 \cdot \mathit{Pr} + \mathit{Se}},
  \nonumber
  \end{equation}
  % \smallskip

\item \emph{F}$^*$: \emph{F-star}:
  \begin{equation}
  \mathit{F}^* = \frac{d}{b + c + d}
  \end{equation}
  \emph{F}$^*$ is a transformation of the F-measure permitting
  more straightforward interpretations (Hand \emph{et
  al.}~\cite{Han21} gives several). It can be alternatively written
  as
  \begin{equation}
  \mathit{F}^* = \frac{F_1}{2 - F_1} \nonumber
  \end{equation}
  or
  \begin{equation}
  \mathit{F}^* = \frac{\mathit{Se} \cdot \mathit{Pr}}
    {\mathit{Se} + \mathit{Pr} - \mathit{Se} \cdot \mathit{Pr}}.
  \nonumber
  \end{equation}
  It is identical to the \emph{Jaccard coefficient} of numerical
  taxonomy and has also been called the \emph{ratio of
  verification}, the \emph{threat score}, the \emph{Tanimoto
  index}, and the \emph{critical success index} in different
  contexts~\cite{Mbi24}.
  \smallskip

\item \emph{FS}: \emph{Symmetric F}~\cite{Sit23}:
  \begin{equation}
  \mathit{FS} = \frac{4}{(b + d)/d + (a + c)/a + (a + b)/a +
    (c + d)/d}
  \end{equation}
  That is,
  \begin{equation}
  \mathit{FS} = \frac{4}{\mathit{Se}^{-1} + \mathit{Sp}^{-1} +
    \mathit{Pr}^{-1} + \mathit{NPV}^{-1}} \nonumber
  \end{equation}
  This is the harmonic mean of \emph{Se}, \emph{Sp}, \emph{PR}, and
  \emph{NPV}, and was developed as a symmetric alternative to the
  F-measure. Note that it can also be written as the harmonic mean
  of \emph{F}$_1$ and \emph{F}', where \emph{F}$_1$ is \emph{F}'
  with the class labels transposed:
  \begin{equation}
  \mathit{FS} = 2 \left( \frac{\mathit{Se}^{-1} + \mathit{Pr}^{-1}}
    {2} + \frac{\mathit{Sp}^{-1} + \mathit{NPV}^{-1}}{2}
    \right)^{- 1} \nonumber
  \end{equation}
% \smallskip

\item \emph{FA}: \emph{Average F-measure}~\cite{Han12}:
  \begin{equation}
  \mathit{FA} = \frac{1}{2} \left[ 2\left( \frac{b + d}{d} +
    \frac{c + d}{d} \right)^{-1} + 2\left( \frac{a + c}{a} +
    \frac{a + b}{a} \right)^{- 1} \right]
  \end{equation}
  That is,
  \begin{equation}
  \mathit{FA} = \frac{1}{2} \left( \frac{2}{\mathit{Se}^{-1} +
    \mathit{Pr}^{-1}} + \frac{2}{\mathit{Sp}^{-1} +
    \mathit{NPV}^{-1}} \right) \nonumber
  \end{equation}
  This is analogous to \emph{FS}, but instead of taking the
  \emph{harmonic} mean of \emph{F}$_1$ and \emph{F}', it takes the
  \emph{arithmetic} mean.
  \smallskip

\item \emph{MCC}: \emph{Matthews Correlation
  Coefficient}~\cite{Mat75,Chi20}:
  \begin{equation}
  \mathit{MCC} = \frac{a \cdot d - b \cdot c}
    {\sqrt{(c + d) (b + d)(a + c)(a + b)}}
  \end{equation}
  Also called the \emph{Phi coefficient} ($\phi$-coefficient),
  \emph{Cram\'er's V}, and the \emph{mean square contingency
  coefficient}. It can also be written as
  \begin{equation}
  \mathit{MCC} = \frac{a \cdot d - b \cdot c}{n^2
    \sqrt{\pi_0\pi_1p_0p_1}} \nonumber
  \end{equation}
  Regarding the true class labels as binary, 0 and 1, and the
  predicted class labels as binary, 0 and 1, this is the Pearson
  product moment correlation between the true and predicted values
  for the test data set.
  \smallskip

\item \emph{PLR}: \emph{Positive Likelihood Ratio}:
  \begin{equation}
  \mathit{PLR} = \left( \frac{d}{b + d} \right) / \left(
    \frac{c}{a + c} \right)
  \end{equation}
  That is
  \begin{equation}
  \mathit{PLR} = \frac{\mathit{Se}}{1 - \mathit{Sp}} \nonumber
  \end{equation}
  or the ratio of the proportion of positive test results amongst
  the positives, to the proportion of positive test results amongst
  the negatives.
  \smallskip

\item \emph{NLR}: \emph{Negative Likelihood Ratio}:
  \begin{equation}
  \mathit{NLR} = \left( \frac{b}{b + d} \right) / \left(
    \frac{a}{a + c} \right)
  \end{equation}
  That is
  \begin{equation}
  \mathit{NLR} = \frac{1 - \mathit{Se}}{\mathit{Sp}} \nonumber
  \end{equation}
  or the proportion of negative test results amongst the positives,
  to the proportion of negative test results amongst the negatives.
  \smallskip

\item \emph{DOR}: \emph{Diagnostic Odds Ratio}~\cite{Gla03}:
  \begin{equation}
  \mathit{DOR} = \left( \frac{d}{b} \right) / \left( \frac{c}{a}
  \right) = \frac{a \cdot d}{b \cdot c}
  \end{equation}
  This is the ratio of the odds of predicting class 1 for class 1
  to the odds of predicting class 1 for class 0. It can also be
  written as
  \begin{equation}
  \mathit{DOR} = \left( \frac{\mathit{Se}}{1 - \mathit{Se}} \right)
    / \left( \frac{1 - \mathit{Sp}}{\mathit{Sp}} \right) =
    \frac{PLR}{NLR} \nonumber
  \end{equation}
  and as
  \begin{equation}
  \mathit{DOR} = \left( \frac{\mathit{Pr}}{1 - \mathit{Pr}} \right)
    / \left( \frac{1 - \mathit{NPV}}{\mathit{NPV}} \right)
    \nonumber
  \end{equation}
  Note that it is independent of class sizes.
  \smallskip

\item \emph{FM}: \emph{Fowlkes-Mallows Index}:
  \begin{equation}
  \mathit{FM} = \sqrt{\left( \frac{d}{b + d} \right) \left(
    \frac{d}{c + d} \right)}
  \end{equation}
  That is, the geometric mean of sensitivity (recall) and precision
  \begin{equation}
  \mathit{FM} = \sqrt{\mathit{Se}\cdot\mathit{\Pr}} \nonumber
  \end{equation}
%  \smallskip

\item \emph{WRACC}: \emph{Weighted Relative
  Accuracy}~\cite{Lav99,Pow11}:
  \begin{equation}
  \mathit{WRACC} = 4 \left( \frac{d}{b + d} - \frac{c + d}{n}
    \right) \left( \frac{b + d}{n} \right)
  \end{equation}
  That is
  \begin{equation}
  \mathit{WRACC} = 4 (\mathit{Se} - p_1) \pi_1 \nonumber
  \end{equation}
  so that it is a rescaled comparison of sensitivity with the
  overall proportion classified as class 1.
  \smallskip

\item \emph{MI}: \emph{Mutual Information}~\cite{Bal00}:
  Letting $a' = a/n$, $b' = b/n$, $c' = c/n$, and $d' = d/n$, we
  have
  \begin{eqnarray}
  \mathit{MI} & = & a' log(a') + b' log(b') + c' log(c') +
    d' log (d') - \\
    & ~ & a' log\left( (a' + b')(a' + c') \right) -
          b' log\left( (a' + b')(b' + d') \right) - \nonumber \\
    & ~ & c' log\left( (a' + c')(c' + d') \right) -
          d' log\left( (d' + b')(d' + c') \right) \nonumber 
  \end{eqnarray}
  This can be interpreted as the reduction in uncertainty about the
  true classes given the predictions.
  
\end{itemize}

%  - - - - - - - - - - - - - - - - - - - - - - - - - - - - - - - - - -

\medskip
\noindent
\textbf{Other measures}:
There are, of course, other, less widely used measures. These have
often been developed for specific applications. One example is
$T_1$~\cite{Han08}, developed for detecting fraudulent plastic card
transactions (e.g.~for credit cards). This is defined as
\begin{equation}
  T_1 = (b \cdot k + c + d) \theta
\end{equation}
where $\theta$ is the cost arising from predicting a transaction as
fraudulent (e.g. the cost of the subsequent investigation), and $k$
is the ratio of the cost of misclassifying a fraudulent transaction
as legitimate to the cost of predicting a transaction as fraudulent.
Details are given in Hand \emph{et al.} (2008)~\cite{Han08}.

Another example is the \emph{Proportion of Explained Variation}
(\emph{PEV})~\cite{Gu09}. In terms of our cell count notation, the
variance of the true class label is $(a + c)(b + d) /n^2$, and the
average variance of the true class label over the two levels of the
predicted class is
\begin{eqnarray}
  \mathit{PEV} &=& \left( \frac{a + b}{n} \right)
    \left( \frac{a}{a + b} \right) \left( \frac{b}{a + b} \right)
    + \left( \frac{c + d}{n} \right) \left( \frac{c}{c + d}
    \right) \left( \frac{d}{c + d} \right) \\ \nonumber
    & = & \frac{1}{n}
    \left( \frac{a \cdot b}{a + b} + \frac{c \cdot d}{c + d} \right)
\end{eqnarray}
The proportion of variance explained by the classifier is thus
\begin{equation}
  n \left( \frac{a \cdot b}{a + b} + \frac{c \cdot d}{c + d} \right)
   / \left( a + c \right) \left( b + d \right). \nonumber
\end{equation}

\begin{table}[t]
  \centering
    \caption{Properties of performance measures. An \textsf{X} means
    the measure satisfies the property. The final column is the
    complement of the ``Complete'' column for the reason explained
    at the end of Section~\ref{sec:application}: neither the
    presence nor the absence of an \textsf{X} is necessarily a
    ``good'' thing.}  \label{tab:properties}
    \begin{small}
    \begin{tabular}{@{}lcccccc@{}} \toprule
    \textbf{Measure}  & \multicolumn{6}{c}{\textbf{Property}} \\
    ~ & Costs & Complete & Symmetry & Meaning & Balanced & Ignore 
      some cells \\
      \midrule
    \emph{Se / R} & & & & \textsf{X} & & \textsf{X} \\
    \emph{Sp} & & & & \textsf{X} & & \textsf{X} \\
    \emph{Pr / PPV} & & & & \textsf{X} & & \textsf{X} \\
    \emph{FDR} & & & & \textsf{X} & & \textsf{X} \\
    \emph{NPV} & & & & \textsf{X} & & \textsf{X} \\
    \emph{FOR} & & & & \textsf{X} & & \textsf{X} \\
    \emph{ACC} & & \textsf{X} & \textsf{X} & \textsf{X} & & \\
    \emph{BACC} & & \textsf{X} & & \textsf{X} & \textsf{X} & \\
    \emph{GACC} & & \textsf{X} & & \textsf{X} & & \\
    \emph{ER} & & \textsf{X} & \textsf{X} & \textsf{X} & & \\
    \emph{WER(k)} & \textsf{X} & \textsf{X} & & \textsf{X} & & \\
    \emph{K} & & \textsf{X} & \textsf{X} & \textsf{X} & & \\
    \emph{J / INF} & & \textsf{X} & & & \textsf{X} & \\
    \emph{MAR} & & \textsf{X} & & & \textsf{X} & \\
    \emph{F}$_1$ & & & & & & \textsf{X} \\
    \emph{F}$_\beta$ & \textsf{X} & & & & & \textsf{X} \\
    \emph{F}$^*$ & & & & & & \textsf{X} \\
    \emph{FS} & & & \textsf{X} & & & \textsf{X} \\
    \emph{FA} & & & \textsf{X} & & & \textsf{X} \\
    \emph{MCC} & & \textsf{X} & & & & \\
    \emph{PLR} & & \textsf{X} & & & & \\
    \emph{NLR} & & \textsf{X} & & & & \\
    \emph{DOR} & & \textsf{X} & & & & \\
    \emph{FM} & & \textsf{X} & & \textsf{X} & & \\
    \emph{WRACC} & & \textsf{X} & & & & \\
    \emph{MI} & & \textsf{X} & \textsf{X} & & & \\
    \bottomrule
    \end{tabular}
    \end{small}
\end{table}

Measures based on amount of variance explained are widely used in
other contexts (e.g. regression), but generally have less relevance
to classification problems. Other measures are based on similar
principles, but using other summary statistics of performance. For
example, in the context of inducing rule sets in knowledge
discovery, Lavrač \emph{et al.} (1999)~\cite{Lav99} describe
several such measures of how the classifier improves prediction
over the baseline proportions in each class, including \emph{WRACC}
mentioned above. Likewise, other methods are based on entropy and
information theory (see, for example, Baldi \emph{et al.}
(2000)~\cite{Bal00} and Valverde-Albacete and
Peláez-Moreno (2020)~\cite{Val20}).

In some cases, the true class sizes will be unknown. For example,
if the classifier is to be applied to future populations where the
relative proportions are likely to differ from those in the training
and test set. In such cases the confusion matrix cannot be reduced
to a single number. There are various ways one can proceed: (i) give
a range of values for the chosen performance measure, corresponding
to different ratios of class sizes (e.g. in a ROC curve), (ii)
average over a distribution of class sizes (e.g. in AUC or the
H-measure -- see Hilden~\cite{Hil91} and Hand and
Anagnostopoulos~\cite{Han22}), (iii) fix one of the degrees of
freedom and optimise the other (e.g. set specificity to 0.9 and
measure the corresponding sensitivity).

It is important to note that a common mistake in evaluating
classification methods is to take a performance measure developed
for a particular application and then use it more widely, ignoring
the special properties of the situation for which it was developed.
The F-measure is often used in this way, in applications far removed
from information retrieval where it was originally developed (for a
discussion see Christen \emph{et al.}~\cite{Chr23}).

% --------------------------------------------------------------------

\section{Related work}
\label{sec:related}

This section reviews reviews of classification performance measures.
The literature is now extensive, so we apologise in advance to
those authors whose work we have missed, and also apologise for the
very few sentences we can devote to summarising the content of each
review we have \emph{not} missed. However, it has to be said that
the literature is also moderately repetitive, with a substantial
proportion of the papers presenting no new perspective on the choice
of methods. On the other hand, we note that perhaps there is some
justification for the repetition on the grounds that researchers
do not have time to read beyond their disciplines, so that
repeating the same ideas in different domains can be very valuable.

We know of several books reviewing and comparing performance
measures (including Hand, 1997~\cite{Han97}; Zhou \emph{et al.},
2014~\cite{Zho14}; Pepe, 2003~\cite{Pep03}; Krzanowski
and Hand, 2009~\cite{Krz09}; Japkowicz and Shah, 2011~\cite{Jap11};
Zou \emph{et al.}, 2011~\cite{Zou11}; and Broemeling,
2012~\cite{Bro12}). Furthermore, when comparing performance measures
it is a not uncommon practice in the machine learning and
statistical literature to present results using more than one
measure, so there is a vast number of implicit comparisons of
measures. We remind the reader, however, that in this paper we are
not concerned with experimental comparisons of performance measures.
This is simply because we consider them to be of limited value.
Unlike comparisons of classification methods themselves, where one
might explore their relative performance on a variety of real data
sets, since different performance measures measure different things
it is not clear what benefit emerges from empirical comparisons of
performance measures -- with the possible exception of explorations
of statistical properties of measures (such as their coefficients
of variation to see which is more discerning).

A paper very much in the spirit of the present work is that by
Kalousis and Theoharis (1999)~\cite{Kal99}, who develop an
intelligent assistant to guide in the choice of classification
method. The main difference is that as well as considering
``accuracy'' performance measures (the kind considered in this
paper) they also consider ``non-accuracy'' measures of the kind we
listed in the introduction, such as execution time, training time,
and resource demand. We have restricted ourselves to the first type
because we believe they are more fundamental -- properties such as
execution time are irrelevant if one is using the wrong measure of
performance.

Baldi \emph{et al.}~(2000)~\cite{Bal00} describe various measures,
including the mutual information discussed in
Section~\ref{sec:measures} -- the reduction in uncertainty of one
variable due to observing the other. Measures of this sort are
valuable in various contexts, but we think less so in contexts
where explicit classification is the aim.

Joshi (2002)~\cite{Jos02} remarks that ``the definition of [an]
effective classifier, embodied in the classifier evaluation metric,
is however very subjective, dependent on the application domain'',
which aligns precisely with our perspective. They compare a range
of measures for problems in which one class is much larger than the
other.

Gail and Pfeiffer (2005)~\cite{Gai05}, in the biomedical context,
review a variety of classification performance criteria,
contrasting some measures with those based on misclassification
costs. They include a discussion of calibration, proportion of
variation explained, and AUC, issues not considered here.

Sokolova \emph{et al.} (2006)~\cite{Sok06} is close in spirit to
the aim of our paper, although more narrowly focused. For example,
they write ``Our argument focusses on the fact that the last four
measures are suitable for applications where one data class is of
more interest than others, for example, search engines,
information extraction, medical diagnoses. They may be not suitable
if all classes are of interest and yet must be distinguished.''
That is, the authors align with our view that requirements of the
problem are what determine the suitability of a measure.

Gu and Pepe (2009)~\cite{Gu09} review standard performance measures
from the biomedical diagnostic perspective. They also discuss
measures such as the proportion of explained variation.

Sokolova and Lapalme (2009)~\cite{Sok09} are concerned with
``accuracy'' properties of performance measures, and present a
systematic analysis of twenty-four measures in terms of eight
invariances to changes in the underlying confusion matrix (their
discussion includes multi-class and multi-label problems). A
familiar example is the constancy of error rate to the distribution
of counts between class 0 points misclassified as class 1 and vice
versa, provided their sum is fixed. Invariances define the
isoeffectiveness curves. Note, however, that Sokolova and Lapalme
(2009) comment that ``We show that, in some learning settings, the
correct identification of positive examples may be important
whereas in others, the correct identification of negative examples
or disagreement between data and classifier labels may be more
significant'', which is an example of the rationale underlying our
paper: the nature of the problem should determine what performance
metric properties are relevant.

Choi \emph{et al.} (2010)~\cite{Cho10} review 76 similarity and
distance measures between pairs of binary vectors. However, they
are particularly interested in unsupervised classification, in
which the vectors being compared describe the same sorts of
objects. This is different from our situation, in which the vectors
are the labels of the objects in the test set, one vector
corresponding to the true class labels and the other to the
predicted class labels: our problem has an intrinsic asymmetry.
That means that not all of the measures considered in Choi
\emph{et al.} may be relevant to our problem.

Powers (2011)~\cite{Pow11} describes sensitivity, precision, and
the F-measure as ignoring ``performance in correctly handling
negative examples'' and says that they ``propagate the underlying
marginal prevalences and biases, and they fail to take account [of]
the chance level performance.'' Our perspective is that whether
negative examples should be ignored, whether the underlying marginal
ratios should be propagated, and whether chance performance level
should be ignored depends on one's aims. That is, these are
\emph{properties} rather than \emph{shortcomings} of the measures,
and may or may not be appropriate depending on the research
question.

Hand (2012)~\cite{Han12} describes a range of classification
performance measures and relationships between measures, spanning
both crisp measures and measures which integrate over a range of
threshold values. He writes ``the core desirable property of a
classification rule [is that] it should, in some sense, classify
as many objects as possible correctly. However, since different
metrics are equivalent to interpreting the phrase ``in some sense''
in different ways, one should expect different metrics to yield
different results. One should expect different metrics to order
classifiers differently.'' He also writes ``None of the measures
can be more `right' or `wrong' than others -- they simply measure
different things.'' As we have already seen in this review, this
is something that some authors forget when criticising measures in
an absolute sense rather than in the context of particular uses.

Hernández-Orallo \emph{et al.} (2012)~\cite{Her12} present an
impressive exploration of threshold choice methods, including
fixed, score-uniform, score-driven, rate-driven, and optimal
methods. Their discussion is thus broader than the crisp measures
discussed in our paper, and includes methods that average over
possible thresholds, such as the AUC, as well as methods based on
the estimated probabilities of class memberships, such as the Brier
score. However, their discussion is narrower than ours in that it
focuses on expected misclassification loss.

Parker (2013)~\cite{Par13} reviews seven standard measures of
classification performance. He also carries out an empirical study,
concluding that ``Both our empirical and theoretical results
converge strongly toward one of the newer methods'' and ``First,
always make a sensible assumption about loss and make this
assumption explicit in the measure. Second, use a measure that
rarely disagrees with all other established measures. These two
criteria appear to be well satisfied by the H-measure, and it is
thus recommended for future classifier evaluations.'' Of course,
our view is that the choice of measure should be based on the aims
of the study.

Sebastiani (2015)~\cite{Seb15} explores formal aspects of
performance measures (the structural aspects described in
Section~\ref{sec:structural}), and so has a rather different aim
from our work. They say they ``adopt an \emph{axiomatic} approach,
i.e., one based on arguing in favour of a number of properties
(``axioms'') that an evaluation measure for classification should
intuitively satisfy.'' In contrast, our position is that the
properties that an evaluation measure should satisfy are determined
by the problem and aims. However, we entirely endorse their comment
that the benefit of the axiomatic approach is that it shifts the
discussion from the measures to the properties -- it is simply that
we do not believe that it is necessary to adopt the axiomatic
approach to do this. Sebastiani (2015) shows that a number of
common measures fail to satisfy his chosen axioms, and defines a
new measure (\emph{K}) which is equal to the Youden index,
\emph{J}, when both classes have some members, and is equal to
$(2 \cdot \mathit{Sp} - 1)$ when class 1 is empty and
$(2 \cdot \mathit{Se} - 1)$ when class 0 is empty.

Hossin and Sulaiman (2015)~\cite{Hos15} focus on model selection
and discuss measures that are based on rankings of object scores,
measures based on estimates of class probabilities, as well as
crisp measures.

Jiao and Du (2016)~\cite{Jia16} review performance measures in
machine learning when applied in bioinformatics. They include a
discussion of AUC as well as methods for crisp measures, and also
some description of measures for multiclass problems.

In an impressively comprehensive series of papers, Canbek \emph{et
al.}~\cite{Can17,Can21,Can23} focus on accuracy based performance
metrics ``such as canonical form, geometry, duality,
complementation, dependency, and levelling'', exploring the
mathematical properties of a diverse range of measures, as well as
the relationships between the measures. They include measures which
assign an estimated class 1 probability to each case, rather than
merely a 0 / 1 class label, and also measures which average over
possible choices of threshold. Canbek \emph{et al.}
(2023)~\cite{Can23} illustrate their work with a case study of
Android Mobile-Malware classification. Since our contention is that
different research situations require different properties of the
performance measure, we have avoided the restriction to a single
exemplar problem. They produce an impressive diagram showing the
relationships between measures in these terms\footnote{See also:
\url{https://github.com/gurol/PToPI}}.

Berrar (2018)~\cite{Ber18} also provides a broader comparison,
including not only crisp measures, but also measures which are
based on ranks (no single threshold specified), and measures which
compare the true class with the estimated probability of belonging
to the true class. They also discuss statistical aspects including
significance and confidence intervals.

Dinga \emph{et al.} (2019)~\cite{Din19} review some threshold-based,
rank-based, and probabilistic measures. They thus extend the
comparison beyond crisp measures. As they write: ``no performance
measure is perfect and suitable for all situations and different
performance measures capture different aspects of model predictions.
Thus, a thoughtful choice needs to be made in order to evaluate the
model predictions based on what is important in each specific
situation.'' They also include empirical comparisons.

Starovoitov and Golub (2020)~\cite{Sta20} carry out an analytic and
experimental comparison of 17 classification performance measures.
They concluded that the AUC (using a non-standard definition) and
Youden index are the ``best estimation functions of both balanced
and imbalanced datasets.'' Of course, we would say that the notion
of absolute ``best'' in this context is meaningless, since different
problems have subtly different aims.

Tharwat (2021)~\cite{Tha21} presents a general review of
classification performance measures, including crisp measures and
graphical tools such as ROC, precision-recall, and detection error
trade-off curves.

G\"osgens \emph{et al.} (2021)~\cite{Gos21} list desirable
properties of measures and look at which measures satisfy each of
them. They cover both binary and multiclass measures, but like us
restricts themself to crisp measures. Table 2 of G\"osgens \emph{et
al.} (2021) corresponds to our Table~\ref{tab:properties}, and
shows which measures have which properties. However, our table
differs in several ways from G\"osgens \emph{et al.}, not merely in
our restriction to binary classification methods. Firstly, our
lists of both measures and properties differ from those of G\"osgens
\emph{et al.} Of course, many properties, including others not
considered by either G\"osgens \emph{et al.} or ourselves, could be
relevant for particular special problems. Secondly, G\"osgens
\emph{et al.}~are more interested in formal properties and, indeed,
establish some interesting mathematical relationships between
various measures, including proving an impossibility theorem
showing that certain properties cannot be simultaneously satisfied.

Dyrland \emph{et al.} (2023)~\cite{Dyr23} eloquently criticise
several widely-used measures because they fail to take account of
the costs of the different kinds of (mis)classification, and gives
several nice examples. Of course, this is based on the assumption
that costs capture the aspects of the problem with which one is
concerned.

Schlosser \emph{et al.} (2023)~\cite{Sch23} produce an extensive
list of measures, and list their properties. This could prove to be
a valuable resource, since they remark that: ``As this manuscript
is meant as a continuously consolidated overview, more evaluation
metrics will be added over time''.

Shirdel \emph{et al.} (2024)~\cite{shi24} provide a brief review of
other work on classification performance measures and then give a
detailed analysis of costs (and their complements, rewards) in
constructing measures, deriving a concept they call
\emph{worthiness} which gives ``the minimal change needed in a
confusion matrix for a classifier to be deemed better than another''.
%\medskip

Other comparison papers are empirical, comparing classifier rankings
attained by different methods when applied to data sets. For example,
Ferri \emph{et al.} (2009)~\cite{Fer09}, Choi \emph{et al.}
(2010)~\cite{Cho10}, Liu \emph{et al.} (2014)~\cite{Liu14}, Jones
\emph{et al.} (2015)~\cite{Jon15}, Luque \emph{et al.}
(2019)~\cite{Luq19}, and Chicco \emph{et al.} (2021)~\cite{Chi21},
experimentally evaluate a selection of performance measures by
applying multiple classification methods on a diverse range of data
sets. The obtained performance results are then grouped, either by
applying a hierarchical clustering algorithm or calculating the
correlation between measures, with the aim to identify measures
which produced relatable results. However, as we have noted above,
we are unconvinced of the merit of experimental investigations,
because we believe that the choice of which measure to use should
be made on the basis that a measure captures the aspect of
performance that is of interest to a user, not if a measure
performs similarly to another measure on a(n arbitrary?) selection
of data sets and classification methods. Ferri \emph{et al.}
(2009)~\cite{Fer09} indeed write (on page 28): ``The results show
that most of these metrics really measure different things and in
many situations the choice made with one metric can be different
from the choice made with another. These differences become larger
for multiclass problems, problems with very imbalanced class
distribution and problems with small datasets.''

Since this work is restricted to crisp measures, we have not
included a discussion of the even more extensive literature on
binary class probability estimation. Important papers in this area
include that of Buja \emph{et al.} (2005)~\cite{Buj05} and
Dimitriadis \emph{et al.} (2024)~\cite{Dim24}.

We conclude this review by remarking that \emph{all} performance
measures have received criticism from researchers. From our
perspective, this is hardly surprising: since different measures
will be appropriate for different aims it necessarily means that
other measures, focused on different aspects, will be
\emph{in}appropriate.

% --------------------------------------------------------------------

\section{Conclusion}
\label{sec:conclusions}

Many measures of performance of classification methods have been
defined, and there is an associated extensive literature. Moreover,
this literature is diverse and spread over many application domains,
including statistics, machine learning, artificial intelligence,
medicine, finance, and others. It includes many comparative studies
applied to both real and simulated data, as well as numerous
discussions of the relative merits of the measures. However, most
of this literature ignores an important distinction between
different kinds of properties of the measures, treating structural
properties (such as whether they lie in the interval $[0,1]$ or
take only positive values) as equally important to conceptual
properties (such as the relative severities of different kinds of
misclassification). It is our contention that these latter
properties are crucial to selecting a measure which properly
reflects the aim of the research or the application, and that
choosing a measure without matching its conceptual properties to
the aim can lead to poor choice of classification method -- and
hence to mistaken classifications, with all the consequences that
might entail. We have attempted to identify these conceptual
aspects, and have produced Table~\ref{tab:properties} showing
which of them are possessed by a wide range of performance
measures. We stress here that in any particular application it
could be that possessing or not possessing a property is what is
needed. That is, possessing a property is not necessarily
desirable -- it depends on the aim.

It has become conventional in certain domains to use particular
measures, with little thought being given to whether they are
appropriate. Dyrland \emph{et al.} (2023)~\cite{Dyr23} nicely
illustrate this, with a story involving a factory manager having
to choose between two classification methods and asking whether
the performance measures used to assess them are appropriate.
``The developers assure [the manager] that these metrics are
widely used. The manager \ldots comments, `I don't remember
``widely used'' being a criterion of scientific correctness --
not after Galileo at least','' and decides to use both classifiers
in a comparative experiment showing actual revenue made (the
aspect of performance which is the one relevant to the problem).
The result is that the classifier which performed worse according
to the ``widely-used metrics'' did better. As the manager says
``it is always unwise to trust the recommendations of developers,
unacquainted with the nitty-gritty reality of a business.'' Or
perhaps even worse: as noted by Parker~\cite{Par13}, the abundance
of performance measures means that ``a researcher could engineer
`better performance' by selecting the right performance measure.''
More generally, we conjecture that lack of attention to the
issues discussed in this paper could mean that many of the
comparative studies of relative performance of classification
methods are irrelevant to the problem they purport to be aimed at
or, worse, could even be misleading.

At a high level, the conceptual aspects we have covered describe
such matters as the need to:

\begin{itemize}
\item Clearly articulate the aims and objectives of the
  classification exercise. A general objective of ``to correctly
  classify as many objects as possible'' may miss critically
  important subtleties of the task.
  \smallskip

\item Identify any constraints on the appropriate measures. For
  example, are all cells of the confusion matrix relevant to the
  objective, should the classes be treated symmetrically, is one
  type of misclassification more serious than the other, and so
  on.
  \smallskip

\item Always report the criteria on which the choice of performance
  measure was based.
  \smallskip

\item Explain why and how the chosen measure matches the aims,
  satisfies the constraints, and meets the criteria.
\end{itemize}

Finally, we should note that since this paper is concerned with
conceptual aspects of the measures, we have not discussed a wide
variety of practical issues which are just as important in
arriving at reliable conclusions of the relative merits of
classifiers in any particular application. These include
statistical estimation and comparison of estimates of measures,
including allowing for multiplicity, correlation, test set
sampling scheme, unknown relative class sizes, and other matters.
As with the structural/conceptual distinction, some of these have
been inadequately considered in the literature.

% ====================================================================

\iffalse % ***********************************************************

\backmatter

\section*{Declarations}

%\bmhead{Acknowledgements}
%
%S. Ziyad likes to acknowledge the Australian Government Research
%Training Program.

\bmhead{Author contributions}

Conception and design: DH and PC; implementation: SZ and PC;
writing: DH and PC; proofreading: DH, PC and SZ.

\bmhead{Funding} PC is funded by the UK Economic and Social
Research Council (ESRC) through project ES/W010321/1; and SZ is
funded by the Australian Government Research Training Program.

\bmhead{Code} Python programs for Examples 4 and 5 in
Section~\ref{sec:intro} are available from: 
\url{https://github.com/SumayyaZiyad/perf-measure}.

\bmhead{Conflict of interest} The authors declare that they have
no conflict of interest relating to the content of this article.

\bmhead{Ethics approval} This article does not contain any
studies with human participants or animals performed by any
of the authors.

\fi % ****************************************************************

\iffalse  % The following needs to be done ***************************

Statements and Declarations

The following statements should be included under the heading "Statements and Declarations" for inclusion in the published paper. Please note that submissions that do not include relevant declarations will be returned as incomplete.

    Competing Interests: Authors are required to disclose financial or non-financial interests that are directly or indirectly related to the work submitted for publication. Please refer to “Competing Interests and Funding” below for more information on how to complete this section.

\section*{Declarations}

Some journals require declarations to be submitted in a standardised format. Please check the Instructions for Authors of the journal to which you are submitting to see if you need to complete this section. If yes, your manuscript must contain the following sections under the heading `Declarations':

\begin{itemize}
\item Funding
\item Conflict of interest/Competing interests (check journal-specific guidelines for which heading to use)
\item Ethics approval and consent to participate
\item Consent for publication
\item Data availability 
\item Materials availability
\item Code availability 
\item Author contribution
\end{itemize}

\noindent
If any of the sections are not relevant to your manuscript, please include the heading and write `Not applicable' for that section. 

\fi % ****************************************************************

\iffalse % ***********************************************************

\begin{appendices}

\section{Section title of first appendix}\label{secA1}

An appendix contains supplementary information that is not an essential part of the text itself but which may be helpful in providing a more comprehensive understanding of the research problem or it is information that is too cumbersome to be included in the body of the paper.




\end{appendices}

\fi % ****************************************************************

% ====================================================================

%\bibliographystyle{spbasic}
%\vskip 0.2in

\bibliographystyle{acm}

\bibliography{paper}

\end{document}